\documentclass[conference]{IEEEtran}
\IEEEoverridecommandlockouts
\usepackage{cite}
\usepackage{amsmath,amssymb,amsfonts}
\usepackage{algorithmic}
\usepackage{graphicx}
\usepackage{url}
\usepackage{textcomp}
\usepackage{multirow}
\usepackage{booktabs}
\usepackage{cite}
\usepackage{microtype}
\usepackage{xcolor}
\def\BibTeX{{\rm B\kern-.05em{\sc i\kern-.025em b}\kern-.08em
    T\kern-.1667em\lower.7ex\hbox{E}\kern-.125emX}}

\usepackage{todonotes}

\usepackage[bookmarks=false,hidelinks]{hyperref}

\def\BibTeX{{\rm B\kern-.05em{\sc i\kern-.025em b}\kern-.08em
    T\kern-.1667em\lower.7ex\hbox{E}\kern-.125emX}}

\begin{document}

\title{Enhancing LLM-based Autonomous Driving with Modular Traffic Light and Sign Recognition}

\author{

Fabian Schmidt$^{1,2}$,
Noushiq Mohammed Kayilan Abdul Nazar$^{1}$,
Markus Enzweiler$^{1}$, 
and Abhinav Valada$^{2}$

\thanks{$^{1}$ Institute for Intelligent Systems, Esslingen University of Applied Sciences, Germany.}
\thanks{$^{2}$ Department of Computer Science, University of Freiburg, Germany.}
\thanks{This work has been funded by the BMBF under support code 13FH544KB2 (ADRIVE-GPT). The authors alone are responsible for the content of the paper. The authors thank the DACHS data analysis cluster, hosted at Esslingen University and co-funded by the MWK within the DFG's "Großgeräte der Länder" program, for providing the computational resources necessary for this research.}
}

\maketitle

\begin{abstract}
Large Language Models (LLMs) are increasingly used for decision-making and planning in autonomous driving, showing promising reasoning capabilities and potential to generalize across diverse traffic situations. However, current LLM-based driving agents lack explicit mechanisms to enforce traffic rules and often struggle to reliably detect small, safety-critical objects such as traffic lights and signs. To address this limitation, we introduce TLS-Assist, a modular redundancy layer that augments LLM-based autonomous driving agents with explicit traffic light and sign recognition. TLS-Assist converts detections into structured natural language messages that are injected into the LLM input, enforcing explicit attention to safety-critical cues. The framework is plug-and-play, model-agnostic, and supports both single-view and multi-view camera setups. We evaluate TLS-Assist in a closed-loop setup on the LangAuto benchmark in CARLA. The results demonstrate relative driving performance improvements of up to 14\% over LMDrive and 7\% over BEVDriver, while consistently reducing traffic light and sign infractions. We publicly release the code and models on \url{https://github.com/iis-esslingen/TLS-Assist}.
\end{abstract}

\section{Introduction}

Large Language Models (LLMs) are increasingly explored as decision makers and planners for autonomous driving and robotics~\cite{werby2024hierarchical,honerkamp2024language,chisari2025robotic,mohammadi2025more}, due to their strong reasoning abilities and, when combined with dedicated encoders, their capacity to process multimodal information. Beyond navigation, language-aware agents offer a key advantage: their inherent world knowledge and reasoning capabilities can improve robustness in long-tail or unforeseen scenarios, where traditional perception-planning pipelines often struggle~\cite{yang2023llm4drive}. Recent frameworks demonstrate that LLM-based autonomous driving agents can leverage these strengths to achieve competitive driving performance~\cite{chen2025end}.

Early LLM-based approaches to autonomous driving often followed a staged pipeline, where perception outputs and navigation cues were first translated into textual scene descriptions that an LLM then mapped to driving actions~\cite{mao2023gpt, wen2024dilu, mei2024leapad}. While this pipeline demonstrated the feasibility of language-based driving, it suffered from error accumulation and limited scalability, as inaccuracies in early perception stages directly propagated into planning. More recent work shifts toward multimodal LLMs (MLLMs) that rely on implicit visual representations extracted from vision encoders, such as visual tokens~\cite{shao2024lmdrive, winter2025bevdriver, renz2025simlingo}. Although these end-to-end capable systems reason more broadly about traffic situations, they still lack precise detection and consistent attention to safety-critical cues. Even advanced MLLMs struggle with fine-grained spatial localization~\cite{guo2024drivemllm, xie2025vlms}, reliable recognition of small-scale objects such as traffic participants~\cite{xing2025openemma}, or traffic lights and signs~\cite{shao2024lmdrive, winter2025bevdriver}. OpenEMMA~\cite{xing2025openemma} highlights the benefits of integrating a specialized object detection module as a ``visual specialist'' to improve reasoning over traffic participants. However, no current approach explicitly enforces compliance with traffic rules such as traffic lights and signs, where a single missed detection can cause a critical safety violation~\cite{pavlitska2023tlr_survey}.

\begin{figure}[t!]
\centering
\includegraphics[width=\linewidth]{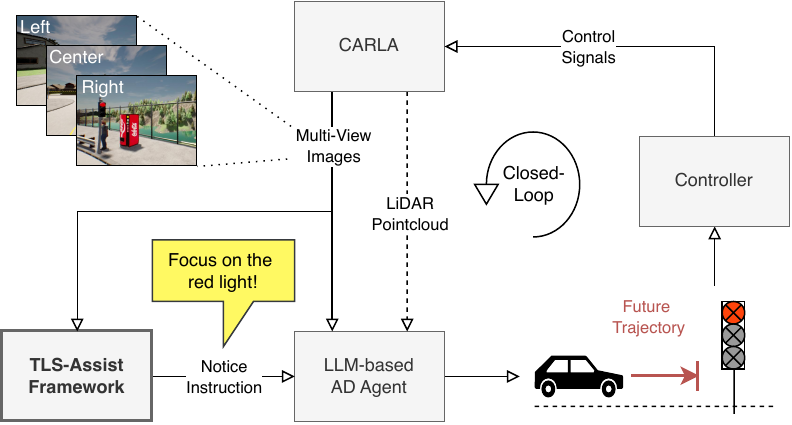}
\caption{Overview of the proposed TLS-Assist framework. TLS-Assist enhances vision-based LLM autonomous driving (AD) agents, with optional LiDAR input, by detecting traffic lights and signs and expressing them as natural language instructions. The extended agents are subsequently evaluated in closed-loop driving using the CARLA simulator.}
\label{fig:tlsr_framework_short}
\end{figure}

To address this gap, we introduce TLS-Assist, a modular framework that augments LLM-based autonomous driving agents with explicit traffic light and sign recognition as a redundancy layer, as illustrated in Fig.~\ref{fig:tlsr_framework_short}. TLS-Assist demonstrates that even lightweight perception priors can substantially enhance closed-loop driving by ensuring explicit attention to traffic rules. The framework operates as a plug-and-play extension compatible with any vision-based LLM stack, supporting both single-view and multi-view camera setups, and converts detections into concise natural language messages that align with the agent’s input. We validate TLS-Assist in CARLA~\cite{dosovitskiy2017carla} on the LangAuto \cite{shao2024lmdrive} benchmark with LMDrive~\cite{shao2024lmdrive} and BEVDriver~\cite{winter2025bevdriver} as baselines, showing consistent improvements in driving performance.

% Our main contributions are:
% \begin{itemize}
%     \item TLS-Assist framework: a modular extension for LLM-based driving agents that provides explicit safety-critical perception priors by detecting traffic lights and signs, applicable to any vision-based LLM AD agent.
%     \item Empirical validation: experiments show that TLS-Assist consistently improves driving performance when integrated into LMDrive and BEVDriver.
%     \item Open-source: trained models and code are made publicly available to foster reproducibility and further research.
% \end{itemize}

To this end, our main contribution is TLS-Assist, a modular extension for LLM-based driving agents that provides explicit safety-critical perception priors by detecting traffic lights and signs. We demonstrate that integrating TLS-Assist into two state-of-the-art language-based driving systems, e.g., LMDrive and BEVDriver, consistently improves closed-loop performance on the LangAuto benchmark. To support reproducibility and future research, we make the code and trained models publicly available.

\section{Related Work}
We review related work on traffic light recognition, traffic sign recognition, and LLM-based autonomous driving.

\subsection{Traffic Light Recognition}
Accurate traffic light recognition (TLR) is a core requirement for autonomous driving, encompassing both detecting the presence and state of lights and determining which of them are relevant to the ego vehicle.

For recognition, early deep learning-based approaches adapted generic object detectors such as YOLO~\cite{redmon2016yolo} or Faster R-CNN~\cite{girshick2015fast} to localize lights and classify their states~\cite{pavlitska2023tlr_survey}. Building on this foundation, recent works emphasize robustness in complex conditions. Polley~\textit{et~al.}~\cite{polley2024tld_ready} combine traffic light detection with road arrow recognition to also handle pictograms, while Chen~\textit{et~al.}~\cite{chen2024traffic} reformulate the task by directly recognizing individual signal bulbs using a two-stage ensemble with data augmentation, improving robustness to varying light arrangements across regions. Tang~\textit{et~al.}~\cite{tang2025real} further enhance recognition of small-scale lights by leveraging RT-DETRv3~\cite{zhao2024detrs}, a real-time transformer-based object detector. Temporal information has also been exploited, with sequence-based models improving state classification across multiple frames~\cite{fan2025vitlr}.

For relevance estimation, a common strategy is to project known 3D traffic light positions from HD maps into the image space~\cite{possatti2019tlr_dl,wu2023autolights}. While effective in pre-mapped environments, these approaches require detailed prior maps and suffer from projection inaccuracies due to localization errors. To avoid reliance on maps, heuristic rules have been applied, e.g., selecting the largest and topmost detected traffic light as relevant~\cite{li2017traffic}, though such assumptions often fail in complex intersections. More advanced strategies fuse multiple cues: Langenberg~\textit{et~al.}~\cite{langenberg2019deep}, for example, combine traffic light, lane marking, and arrow positions in a deep learning-based framework, though their reliance on human-annotated ground truth inputs limits real-world applicability.

\subsection{Traffic Sign Recognition}
Traffic sign recognition (TSR) is equally critical for autonomous driving, providing regulatory and safety information. Although signs are typically closer to the ego lane than traffic lights, their small size and high environmental variability pose significant detection challenges.

YOLO-based adaptations dominate recent research \cite{chen2025yolo_ts, lin2025yolo_llts, sun2024llth, yu2022traffic}. YOLO-TS~\cite{chen2025yolo_ts} improves small-object recognition by adjusting feature resolutions and receptive fields. YOLO-LLTS~\cite{lin2025yolo_llts} and LLTH-YOLOv5~\cite{sun2024llth} extend these efforts to low-light conditions: YOLO-LLTS enhances degraded images to improve reliability under noise, blur, and poor contrast, while LLTH-YOLOv5 integrates a dedicated low-light enhancement stage, significantly improving small-target detection accuracy while maintaining real-time performance.
Beyond single-frame detection, Yu~\textit{et~al.}~\cite{yu2022traffic} propose a multi-image fusion model that combines YOLOv3~\cite{redmon2018yolov3} detection with VGG19~\cite{simonyan2014vgg} recognition across consecutive frames, exploiting temporal redundancy for robustness against occlusion and distance. Wang~\textit{et~al.}~\cite{wang2023vehicle} introduce an adaptive traffic sign detector with a lightweight feature extractor and an adaptive enhancement network that dynamically adjusts to fog, blur, or low illumination, achieving both higher accuracy and real-time performance on embedded platforms.

\subsection{LLM-based Autonomous Driving}
Recent work explores LLMs as decision-makers and planners for driving. Approaches can be distinguished by the form of input provided to the LLM: implicit feature representations in the form of visual tokens~\cite{shao2024lmdrive, winter2025bevdriver, renz2025simlingo}, explicit semantic information such as textual scene descriptions~\cite{mei2024leapad}, or structured object detections~\cite{xing2025openemma}.

LMDrive~\cite{shao2024lmdrive} introduces a language-guided, end-to-end stack that fuses multi-modal sensors (cameras and LiDAR) with natural-language navigation and evaluates in closed loop on CARLA.
BEVDriver~\cite{winter2025bevdriver} extends LMDrive and integrates a more sophisticated multi-modal vision encoder to produce a richer BEV feature map, improving spatial grounding and robustness.
SimLingo~\cite{renz2025simlingo} is a vision-only VLM-based system that jointly targets closed-loop driving and language-action alignment, ensuring that linguistic responses are consistent with executed actions.
In contrast, LeapAD~\cite{mei2024leapad} adopts a dual-process design and employs a VLM to describe the current scene, passing the resulting textual description to an LLM that selects an appropriate driving action.
OpenEMMA~\cite{xing2025openemma} augments the LLM with a dedicated object detection specialist, providing 3D bounding boxes of surrounding traffic participants. These detections are transformed into structured descriptions that are incorporated into the LLM’s reasoning.

Similar to the idea of OpenEMMA, our TLS-Assist module integrates an external perception component, but with a focus on \emph{safety-critical perception priors}: it detects traffic lights and signs and injects structured messages into LLM-based autonomous driving agents. This module provides explicit traffic-rule awareness that current LLM stacks typically lack, aiming for more safety-informed decision-making and planning.

\section{Method}

\begin{figure*}[h!]
\centering
\includegraphics[width=\linewidth]{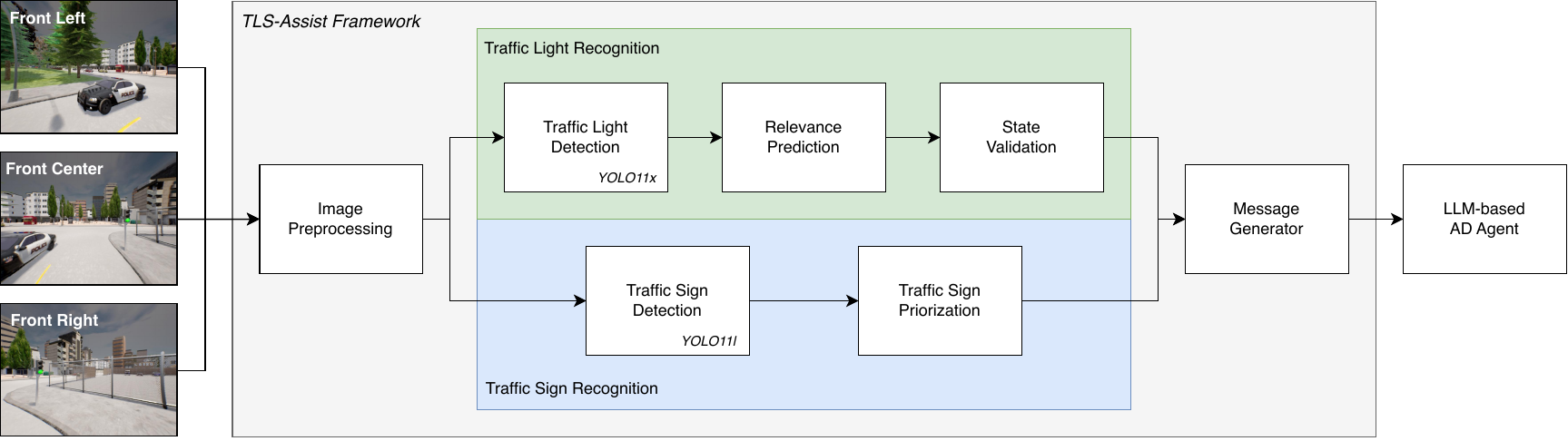}
\caption{Overview of the proposed TLS-Assist framework. Multi-view images are preprocessed and passed to separate detection modules for traffic lights and traffic signs. Traffic light detections are refined by relevance prediction and state validation, while traffic sign detections are prioritized according to driving relevance. Both results are consolidated into natural language messages and forwarded to the LLM-based autonomous driving (AD) agent.}
\label{fig:tlsr_framework}
\end{figure*}

% \subsection{Traffic Light and Sign Recognition Framework}

We propose TLS-Assist, a traffic light and sign assistance framework that provides LLM-based autonomous driving agents with explicit traffic-rule awareness.
The framework integrates with the agent as a modular extension that outputs natural language messages describing the current traffic context. As illustrated in Fig.~\ref{fig:tlsr_framework}, TLS-Assist consists of seven key components: image preprocessing, traffic light detection, relevance prediction, state validation, traffic sign detection, traffic sign prioritization, and a message generator.

The workflow of the framework begins with multi-view image preprocessing, followed by parallel detection of traffic lights and signs. Traffic light detections are refined through relevance prediction and state validation to ensure that only the most critical state is passed to the agent. Traffic sign detections are filtered and prioritized before being translated into messages. Finally, the message generator combines validated traffic light states and traffic sign information into structured natural language instructions, which are injected into the textual input of the autonomous driving agent.
We discuss these components in detail in the following subsections.

\subsection{Image Preprocessing}

The preprocessing module supports both single- and multi-view camera configurations. In the multi-view setup, three RGB cameras (front-left, front-center, and front-right) capture complementary perspectives, whereas in the single-view setup, only the front-center camera is used. All available views are standardized for format and consistency and then stitched into a single panoramic image. This unified representation allows the recognition modules to process one consolidated input per frame, improving efficiency by avoiding duplicate analysis of overlapping regions.

The stitched image is subsequently cropped to predefined field-of-view dimensions of $720px \times 1280px$, before being passed to the recognition modules. This step reduces computational overhead while retaining all relevant visual content. In cases where stitching fails, the system reverts to the single front-center view to ensure uninterrupted operation.

\subsection{Traffic Light Recognition}

Traffic light recognition is formulated as a multi-class object detection task with the categories \textit{red}, \textit{yellow}, \textit{green}, and \textit{off}. The outputs are filtered by a confidence threshold and represented as bounding boxes with associated class labels.  

The \textit{relevance prediction} module resolves ambiguities when multiple traffic lights are detected in a frame. To identify the most likely relevant state, the module applies a majority-based strategy. If no clear majority exists, a predefined priority order (red $>$ yellow $>$ green $>$ off) is applied to ensure that the most safety-critical option is selected. This lightweight heuristic avoids reliance on high-definition maps or external priors while ensuring that the agent consistently attends to the traffic light most relevant for safe driving.

The \textit{state validation} module improves robustness by enforcing temporal consistency and prioritizing safety-critical states. To address fluctuations caused by occlusions, misclassifications, or confidence drops, it aggregates predictions from a sliding buffer of the current and past $N-1$ frames.
Formally, let $B = \{f_{n-(N-1)}, \dots, f_n\}$ denote the buffer of detected traffic light states. The score for each candidate state $c \in \{\text{red}, \text{green}, \text{yellow}, \text{no\_detection}\}$ is defined as
\begin{equation}
W(c) = \sum_{k=0}^{N-1} (N-k) \cdot S(c) \cdot 1_{{f_{n-k} = c}},
\end{equation}
where $(N-k)$ emphasizes recency and $S(c)$ encodes state criticality ($S(\text{red})=3$, $S(\text{green})=2$, $S(\text{yellow})=1$, $S(\text{no\_detection})=0$). The final state is then selected as
\begin{equation}
\hat{c} = \arg\max_{c} W(c).
\end{equation}

In our implementation, we use $N=3$, i.e., the current frame and two history frames, which provides a good trade-off between stability and responsiveness. This formulation ensures that recent and safety-critical detections dominate the decision, thereby suppressing spurious state changes.

\subsection{Traffic Sign Recognition}

Traffic sign recognition is treated as a multi-class detection task with the categories \textit{stop}, \textit{yield}, \textit{speed limit 30}, \textit{speed limit 60}, \textit{speed limit 90}, and \textit{off}. To avoid redundant information, duplicate detections of the same sign are removed. In addition, very small detections are filtered out to ensure that only signs within a meaningful and actionable distance of the ego vehicle are considered.
The remaining detections are ranked according to their driving relevance. A fixed priority order is applied, giving precedence to critical safety signs (\textit{stop} $>$ \textit{yield}) over regulatory ones (\textit{speed limit 30} $>$ \textit{speed limit 60} $>$ \textit{speed limit 90}), while the absence of any sign is treated with the lowest priority (\textit{no\_detection}). This prioritization ensures that the agent receives concise and safety-critical instructions in situations where multiple signs are visible simultaneously.

\subsection{Message Generation}

The message generator consolidates the validated traffic light state and prioritized traffic sign detections into textual instructions based on predefined templates. Separate dictionaries are maintained for traffic lights and traffic signs, each mapping the detected class to a corresponding natural language message. For example, a red light triggers the message \textit{“Red light ahead, stop the vehicle!”}, while a speed limit sign generates an instruction of the form \textit{“Limit speed to $x$ km/h.”}. In cases where no relevant object is detected, the corresponding dictionary entry is empty to avoid redundant messages.

\begin{figure}[t!]
\centering
\includegraphics[width=0.9\linewidth]{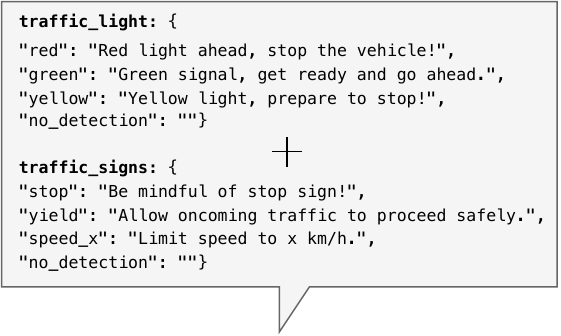}
\caption{Predefined notice instructions used for message generation. Traffic light and sign detections are mapped to natural language templates that are concatenated and provided as textual input to the LLM-based driving agent.}
\label{fig:notice_instructions}
\end{figure}

The final instruction is obtained by concatenating the traffic light message with the prioritized traffic sign message. This ensures that the LLM-based autonomous driving agent receives concise and contextually relevant guidance. An overview of the predefined templates is shown in Fig.~\ref{fig:notice_instructions}, covering all traffic light states and supported traffic sign types.

\subsection{Training Details and Model Selection}

We considered two families of object detection models for traffic light and sign recognition: RT-DETRv3~\cite{wang2025rt_detr3}, a transformer-based detector optimized for real-time performance, and YOLO11~\cite{jocher2024yolo11}, the latest iteration of the YOLO family known for efficiency and strong small-object detection. While RT-DETRv3 demonstrates strong accuracy on COCO~\cite{lin2014coco}, YOLO11 achieves competitive results with fewer parameters, making it computationally attractive. Based on this trade-off, we selected YOLO11 as the backbone for our framework. All models were initialized with COCO-pretrained weights and adapted to our tasks through staged training.

\paragraph{Datasets and Training Procedure.}

For TLR, we adopted a two-stage training procedure. Pre-training was performed for 300 epochs on the ATLAS dataset~\cite{polley2025atlas} (33k images, 73k annotated traffic lights across 25 categories, collected in Karlsruhe, Germany). Fine-tuning was then conducted for 50 epochs on the CARLA traffic light dataset~\cite{affinis2024trafficlightdetection}, which provides 588 annotated images for red and green signals. Due to its small size, we applied six-fold cross-validation, with 83\% of the data used for training and 17\% for validation in each fold.

For TSR, we created a custom dataset tailored to the CARLA environment, since built-in sign classes are limited and differ from real-world distributions. The dataset comprises 4,388 images across five classes (\textit{stop}, \textit{yield}, \textit{speed limit 30}, \textit{speed limit 60}, \textit{speed limit 90}), generated by compositing sign icons onto CARLA background scenes. Augmentation techniques (scaling, flipping, and rotation) were applied during construction. The dataset was split into training, validation, and test sets in an 80:10:10 ratio, and models were trained for 300 epochs.

\paragraph{Model comparison and selection}

We evaluated YOLO11s, YOLO11l, and YOLO11x on both traffic light and traffic sign detection using F1-Score, mAP\textsubscript{50--95}, and inference time, as Precision, Recall, and mAP\textsubscript{50} were saturated across models. Results are shown in Table~\ref{table:model_comparison}. For TLR, YOLO11x achieved the highest mAP\textsubscript{50--95}, demonstrating stronger robustness across IoU thresholds; therefore, we selected it for the traffic light module. For TSR, YOLO11x and YOLO11l performed nearly identically and slightly outperformed YOLO11s, with YOLO11l chosen as the backbone for its higher efficiency.

\begin{table}[!t]
\centering
\caption{Comparison of YOLO11 model variants for Traffic Light (TLR) and Traffic Sign Recognition (TSR). 
We report parameter count in million, F1-Score, mAP\textsubscript{50--90}, and inference time in milliseconds.}
\begin{tabular}{cc cccc}
    \toprule
    & Model & \#Params & F1-Score & mAP\textsubscript{50--95} & Inference [ms] \\
    \midrule
    \multirow{3}{*}{\rotatebox{90}{TLR}} & YOLO11s & 9.4  & 0.928 & 0.869 & 3.32 \\
    & YOLO11l & 26.3 & 0.999 & 0.861 & 7.98 \\
    & YOLO11x & 56.9 & 0.999 & 0.905 & 18.53 \\
    \midrule
    \multirow{3}{*}{\rotatebox{90}{TSR}} & YOLO11s & 9.4  & 0.997 & 0.971 & 2.87 \\
    & YOLO11l & 26.3 & 0.999 & 0.974 & 7.46 \\
    & YOLO11x & 56.9 & 0.998 & 0.974 & 12.93 \\
    \bottomrule
\end{tabular}
\label{table:model_comparison}
\end{table}

\section{Experiments}

This section examines the performance of TLS-Assist and its individual modules (TLR-only and TSR-only), presents an ablation study of TLR’s components, and concludes with a discussion interpreting the results.

\begin{table*}[!ht]
\centering
\caption{Driving Score (DS), Route Completion (RC), and Infraction Score (IS) on LangAuto benchmarks \textit{Tiny}, \textit{Short}, and \textit{Long}. Overall represents the mean across all routes. Best values are highlighted in bold.}
\begin{tabular}{l ccc ccc ccc ccc}
    \toprule
    Method 
    & \multicolumn{3}{c}{Tiny} 
    & \multicolumn{3}{c}{Short} 
    & \multicolumn{3}{c}{Long} 
    & \multicolumn{3}{c}{Overall} \\
    \cmidrule(lr){2-4} \cmidrule(lr){5-7} \cmidrule(lr){8-10} \cmidrule(lr){11-13}
     & DS & RC & IS  
     & DS & RC & IS
     & DS & RC & IS
     & DS & RC & IS \\
    \midrule
    % \multicolumn{13}{l}{\textbf{LMDrive}} \\
    LMDrive \cite{shao2024lmdrive}
      & 52.19 & 62.33 & 0.82
      & 41.50 & 55.63 & 0.81
      & 27.46 & 40.36 & 0.73
      & 40.38 & 52.77 & 0.79 \\
    + TLR-only
      & \textbf{59.87} & \textbf{69.35} & \textbf{0.85}
      & \textbf{47.09} & 57.85 & \textbf{0.84}
      & \textbf{31.28} & \textbf{40.72} & \textbf{0.77}
      & \textbf{46.08} & \textbf{56.64} & \textbf{0.82} \\
    + TSR-only
      & 55.71 & 67.03 & 0.82
      & 46.94 & \textbf{58.55} & 0.83
      & 27.48 & 39.13 & 0.74
      & 43.38 & 54.90 & 0.80 \\
    + TLS-Assist
      & 56.34 & 68.81 & 0.82
      & 44.26 & 57.17 & 0.82
      & 27.11 & 37.88 & \textbf{0.77}
      & 42.57 & 54.62 & 0.80 \\

    \midrule
    
    % \multicolumn{13}{l}{\textbf{BEVDriver}} \\
    BEVDriver \cite{winter2025bevdriver}
      & 63.15 & 68.31 & \textbf{0.92}
      & 42.22 & 46.00 & \textbf{0.92}
      & 28.74 & 34.79 & \textbf{0.86}
      & 44.70 & 49.70 & \textbf{0.90} \\
    + TLR-only 
      & \textbf{65.82} & \textbf{72.75} & 0.88
      & 49.83 & 58.89 & \textbf{0.92}
      & \textbf{30.42} & 37.30 & 0.84
      & \textbf{48.69} & \textbf{55.31} & 0.88 \\
    + TSR-only 
      & 63.57 & 71.44 & 0.88
      & 45.19 & 52.77 & 0.90
      & 30.41 & 36.48 & 0.84
      & 46.39 & 53.56 & 0.87 \\
    + TLS-Assist
      & 62.75 & 71.70 & 0.87
      & \textbf{51.44} & \textbf{60.20} & 0.88
      & 30.16 & \textbf{39.45} & 0.83
      & 48.12 & 57.12 & 0.86 \\
    
    \bottomrule
\end{tabular}
\label{table:results}
\end{table*}

\subsection{Evaluation Setting}

\subsubsection{Benchmark}
We use the LangAuto benchmark~\cite{shao2024lmdrive}, which adapts the CARLA Leaderboard to evaluate language-based driving agents. It spans eight public CARLA towns under diverse traffic layouts, weather, and daylight conditions, and provides natural language navigation instructions instead of discrete commands. LangAuto is divided into three tracks based on route length: \textit{Tiny} ($<150m$), \textit{Short} ($150-500m$), and \textit{Long} ($>500m$).  

\subsubsection{Metrics}
We adopt the official CARLA Leaderboard metrics, which jointly capture driving performance and safety: 

\begin{itemize}
    \item Route Completion (RC): the percentage of the route successfully completed. Deviating too far from the predefined path terminates the episode and penalizes RC.  
    \item Infraction Score (IS): a multiplicative penalty factor that accounts for collisions and traffic rule violations, including ignoring traffic lights or traffic signs, as well as termination causes such as route deviations and route timeouts. The score starts from 1.0 and is decayed according to the type and number of infractions.  
    \item Driving Score (DS): the primary evaluation metric defined for each route $i$ as
    \begin{equation}
        DS_i = RC_i \cdot IS_i ,
    \end{equation}
    combining progress (RC) and safety (IS). Final scores are reported as the arithmetic mean over all routes.  
\end{itemize}

Following the LangAuto evaluation protocol, we repeat each experiment three times and report the average results.

\subsubsection{Baselines}
We use LMDrive~\cite{shao2024lmdrive} and BEVDriver~\cite{winter2025bevdriver} as baselines. Both systems represent state-of-the-art LLM-based autonomous driving approaches, designed to process visual and natural language inputs. Their modular architectures make them compatible with our extension, allowing us to integrate an additional language signal derived from TLS-Assist.  

\subsection{Results}

Table~\ref{table:results} summarizes the performance of TLS-Assist when integrated into LMDrive and BEVDriver.\footnote{We report reproduced results using the official checkpoints rather than the results stated in the paper, since the authors of LMDrive reported that the performance varies significantly depending on GPU model (see issues 37, 52, and 56 in the LMDrive repository). The same behavior was observed for BEVDriver.} Across both baselines, the individual modules (TLR-only and TSR-only) as well as the combined TLS-Assist setup were tested, showing that explicit traffic-rule perception benefits both agents, though in different ways.  

For LMDrive, the TLR-only setting yields the largest gain, raising overall DS from 40.38 to 46.08, with both RC and IS improving. The TSR-only setting also increases performance, but the effect is less pronounced (DS 43.38). Combining both modules into TLS-Assist does not amplify the gains, with DS dropping slightly below the single-module variants. Examining the benchmarks, on \textit{Tiny} routes, RC improves across all variants, while IS is mostly stable except for TLR-only, which shows a clear increase. On \textit{Short} routes, both RC and IS rise consistently, producing the strongest DS gains. In contrast, \textit{Long} routes see lower RC but improved IS partly compensates, resulting in comparable DS and, in the case of TLR-only, even a modest improvement over the baseline.

For BEVDriver, which already achieves stronger baseline performance, TLS-Assist further improves driving behavior. TLR-only raises DS from 44.70 to 48.69, and TLS-Assist achieves a comparable DS of 48.12 but with higher RC. This comes at a trade-off: RC improves consistently, most notably on \textit{Short} routes, where it jumps from 46.00 to 60.20, while IS decreases slightly across all benchmarks. On \textit{Long} routes, RC gains still lift DS moderately despite the IS drop, whereas improvements on \textit{Tiny} routes remain small and inconsistent due to the strong baseline.  

In summary, LMDrive and BEVDriver both benefit from TLS-Assist integration, but in different ways. LMDrive achieves larger relative gains in DS from the traffic-rule modules, particularly TLR-only, though improvements are concentrated on shorter routes and less consistent on long routes. BEVDriver, in contrast, shows smaller absolute gains due to its stronger baseline, but achieves consistent RC improvements across all tracks. Importantly, BEVDriver maintains higher IS throughout, indicating safer driving with fewer infractions, while LMDrive also benefits from TLS-Assist integration by consistently improving both RC and IS when individual modules are added. 

\subsection{Ablation Studies}

\begin{table*}[!ht]
\centering
\caption{Ablation (Abl.) study of TLR components (Traffic Light Detection (D), Relevance Prediction (RP), and State Validation (SV)) on LMDrive and BEVDriver across \textit{Tiny}, \textit{Short}, and \textit{Long} routes of LangAuto benchmark. Overall represents the mean across routes.}
\begin{tabular}{cc ccc ccc ccc ccc ccc}
    \toprule
    & Abl. & D & RP & SV 
         & \multicolumn{3}{c}{Tiny} 
         & \multicolumn{3}{c}{Short} 
         & \multicolumn{3}{c}{Long} 
         & \multicolumn{3}{c}{Overall} \\
    \cmidrule(lr){6-8} \cmidrule(lr){9-11} \cmidrule(lr){12-14} \cmidrule(lr){15-17}
         &   &   &   &   
         & DS & RC & IS 
         & DS & RC & IS 
         & DS & RC & IS 
         & DS & RC & IS \\
    % \midrule
    % \multicolumn{16}{c}{LMDrive} \\
    \midrule
    \multirow{4}{*}{\rotatebox{90}{LMDrive}} & 0 & x & x & x & \textbf{59.87} & \textbf{69.42} & \textbf{0.85} & \textbf{47.09} & 57.85 & \textbf{0.84} & \textbf{31.28} & \textbf{40.72} & \textbf{0.77} & \textbf{46.08} & \textbf{56.00} & \textbf{0.82} \\
    & 1 & x & x &   & 56.81 & 68.59 & 0.82 & 45.64 & \textbf{58.57} & 0.83 & 27.36 & 36.48 & \textbf{0.77} & 43.27 & 54.55 & 0.81 \\
    & 2 & x &   & x & 58.41 & 69.00 & 0.84 & 44.66 & 55.47 & \textbf{0.84} & 26.35 & 36.81 & 0.76 & 43.14 & 53.76 & 0.81 \\
    & 3 & x &   &   & 56.61 & 68.40 & 0.83 & 43.35 & 56.57 & 0.82 & 27.43 & 38.77 & 0.76 & 42.46 & 54.58 & 0.80 \\
    % \midrule
    % \multicolumn{16}{c}{BEVDriver} \\
    \midrule
    \multirow{4}{*}{\rotatebox{90}{BEVDriver}} & 0 & x & x & x & 65.82 & 72.75 & 0.88 & 49.83 & 55.89 & 0.92 & \textbf{30.42} & 37.30 & \textbf{0.84} & \textbf{48.69} & 55.31 & \textbf{0.88} \\
    & 1 & x & x &   & \textbf{65.91} & \textbf{72.80} & \textbf{0.89 }& 47.86 & 55.61 & 0.87 & 27.72 & \textbf{38.02} & 0.76 & 47.16 & \textbf{55.48} & 0.84 \\
    & 2 & x &   & x & 58.43 & 64.66 & \textbf{0.89} & \textbf{54.08} & \textbf{57.00} & \textbf{0.95} & 27.68 & 36.08 & 0.80 & 46.73 & 52.58 & \textbf{0.88} \\
    & 3 & x &   &   & 61.74 & 68.33 & \textbf{0.89} & 49.27 & 55.33 & 0.91 & 29.97 & 36.37 & \textbf{0.84} & 46.99 & 53.34 & \textbf{0.88} \\
    \bottomrule
\end{tabular}
\label{table:ablation_full}
\end{table*}

Table~\ref{table:ablation_full} reports the results of the ablation study on TLR’s internal components: relevance prediction and state validation.  
For LMDrive, the full configuration (ablation 0) achieves the highest overall DS of 46.08. Removing state validation (ablation 1) or relevance prediction (ablation 2) reduces DS to 43.27 and 43.14, respectively, while removing both (ablation 3) yields the lowest performance at 42.46. The drop is most pronounced on \textit{Long} routes, where RC falls from 40.72 to around 36.5, while IS is stable. Across all ablations, IS stays relatively consistent, with the best values observed for the full configuration.  
For BEVDriver, results are more stable overall. The full configuration (ablation 0) achieves a DS of 48.69. Removing state validation (ablation 1) or relevance prediction (ablation 2) slightly reduces DS to 47.16 and 46.73, respectively, while removing both (ablation 3) yields 46.99. RC shows larger fluctuations, especially on \textit{Long} routes, where it drops from 37.30 (full setup) to 28.02 when state validation is removed. IS remains consistently high across ablations, with the highest values observed on \textit{Short} routes.

Overall, LMDrive and BEVDriver achieve comparable levels of RC across ablations, but BEVDriver maintains consistently higher IS, resulting in an improved DS.

\subsection{Discussion}

\begin{table}[!b]
\centering
\setlength{\tabcolsep}{4.75pt}
\caption{Average infractions and termination causes across Tiny, Short, and Long routes of the LangAuto benchmark. Lower values indicate better performance.}
\begin{tabular}{lcccc}
\toprule
Method & Red Light & Stop Sign & Route Dev. & Timeouts \\
\midrule

LMDrive     & 5.86 & 0.37 & 17.13 & 3.23 \\
+ TLR-only       & 5.11 (-13\%) & 0.36 (-4\%)  & 16.34 (-5\%)  & 3.50 (+8\%) \\
+ TSR-only       & 5.78 (-1\%)  & \textbf{0.07} (-81\%) & \textbf{16.11} (-6\%)  & 3.62 (+12\%) \\
+ TLS-Assist      & \textbf{4.22} (-28\%) & \textbf{0.07} (-81\%) & 17.15 (+0\%)  & \textbf{3.17} (-2\%) \\

\midrule

BEVDriver   & 3.74 & 0.74 & 13.66 & 11.87 \\
+ TLR-only       & \textbf{1.34} (-64\%) & 0.71 (-4\%)  & 10.28 (-25\%) & 8.29 (-30\%) \\
+ TSR-only       & 3.46 (-7\%)  & 0.53 (-28\%) &  \textbf{9.91} (-27\%) & 7.70 (-35\%) \\
+ TLS-Assist      & 1.65 (-56\%) & \textbf{0.49} (-33\%) & 12.28 (-10\%) & \textbf{5.84} (-51\%) \\

\bottomrule
\end{tabular}
\label{table:infractions_extended}
\end{table}

% - Traffic Sign improvements not that significant due to less frequent traffic signs in benchmarks.
% - Route Deviation and Timeouts for LMDrive no significant change, but for BEVDriver both metrics are noticeable reduced that leads to an increase of RC. Increased RC also leads to more possible Infractions that explains why IS is lower for TLSR enhanced variants than for Baseline.
% - For both LMDrive and BEVDriver, red light and stop sign infractions are consistently lowered compared to baseline 
% - why does TLSR perform worse than TLR and TSR standalone for LMDrive?

Table~\ref{table:infractions_extended} provides additional insights into infractions and termination causes, complementing the aggregate metrics presented earlier. Several key trends emerge:

The explicit traffic light and sign modules successfully reduce violations for both baselines. Across LMDrive and BEVDriver, red light and stop sign infractions decrease consistently compared to the baseline agents, with the largest reductions observed when both modules are active (TLS-Assist). These improvements highlight the effectiveness of explicitly injecting safety-critical cues. % into the LLM input. 
The gains are more pronounced for red light compliance, while stop sign improvements are less consistent. This result can be explained by the benchmark distribution, where traffic signs are relatively infrequent, reducing their influence on overall behavior.

The impact on route deviation and timeouts differs across the two agents. For LMDrive, these termination causes remain largely unchanged, suggesting that TLS-Assist primarily targets compliance rather than overall navigation robustness. In contrast, BEVDriver shows substantial reductions in both route deviation and timeouts across all module variants, with the largest improvement under TLS-Assist. These reductions directly explain the strong gains in RC observed earlier. However, the higher RC also exposes the agent to longer driving times and, therefore, a greater chance of accumulating infractions. This explains why BEVDriver’s IS decreases slightly despite fewer red light and stop sign violations.

While TLR-only and TSR-only individually improve LMDrive’s performance, their combination into TLS-Assist does not lead to further gains and, in fact, reduces overall DS, at least for LMDrive. An explanation is that combining both modules introduces redundant or conflicting textual cues, which can dilute the ability of the LLM to prioritize relevant signals. This suggests that the integration of structured safety cues must balance informativeness with conciseness to avoid overwhelming the LLM’s reasoning process.

In summary, TLS-Assist enhances traffic rule compliance across both agents, with particularly strong reductions in red light and stop sign violations. BEVDriver benefits most in terms of route completion due to fewer deviations and timeouts, though this comes with a trade-off in IS. LMDrive shows consistent compliance improvements but no systematic gains in navigation robustness, and suffers when both modules are combined. These findings underline that while modular redundancy is effective, the interaction between cues and the LLM’s reasoning pipeline must be carefully managed.

% \subsubsection{Limitations}
% Over simplified assumptions for traffic light recognition (majority voting) and for traffic sign recognition (sign priorization) need to be refined, since they are specifically tailored for the LangAuto benchmark and CARLA environment.

\section{Conclusion}
We introduced TLS-Assist, a modular redundancy layer that augments vision-based LLM autonomous driving agents with explicit traffic light and sign recognition. By converting detections into concise, rule-focused messages, TLS-Assist injects lightweight safety priors into the agent’s reasoning without modifying the planner. In closed-loop evaluation on LangAuto benchmark using CARLA, integrating TLS-Assist yields consistent gains: up to 14.1\% relative improvement in Driving Score for LMDrive and 7.7\% for BEVDriver, alongside marked reductions in red-light (-64\%) and stop-sign (-81\%) violations. For LMDrive, improvements stem from both higher route completion and a consistent reduction in infractions, resulting in an increased infraction score. For BEVDriver, the main gains come from substantially higher route completion due to fewer deviations and timeouts, while the infraction score decreases slightly because longer routes naturally increase exposure to potential infractions.

% \section*{Acknowledgment}
% We thank Sri Harish Madampur Suresh for his support with the evaluations and Sabrina Kaniewski for her valuable assistance with reviewing.

\bibliographystyle{IEEEtran}
\bibliography{bibliography}

@inproceedings{shao2024lmdrive,
    title={Lmdrive: Closed-loop end-to-end driving with large language models},
    author={Shao, Hao and Hu, Yuxuan and Wang, Letian and Song, Guanglu and Waslander, Steven L and Liu, Yu and Li, Hongsheng},
    booktitle={Proceedings of the IEEE/CVF Conference on Computer Vision and Pattern Recognition},
    pages={15120--15130},
    year={2024}
}

@inproceedings{winter2025bevdriver,
    title={BEVDriver: Leveraging BEV Maps in LLMs for Robust Closed-Loop Driving},
    author={Winter, Katharina and Azer, Mark and Flohr, Fabian B},
    booktitle={IEEE/RSJ International Conference on Intelligent Robots and Systems},
    year=2025,
}

@inproceedings{possatti2019tlr_dl,
    title={Traffic light recognition using deep learning and prior maps for autonomous cars},
    author={Possatti, Lucas C and Guidolini, R{\^a}nik and Cardoso, Vinicius B and Berriel, Rodrigo F and Paix{\~a}o, Thiago M and Badue, Claudine and De Souza, Alberto F and Oliveira-Santos, Thiago},
    booktitle={2019 international joint conference on neural networks (IJCNN)},
    pages={1--8},
    year={2019},
    organization={IEEE}
}

@inproceedings{wu2023autolights,
    title={aUToLights: a robust multi-camera traffic light detection and tracking system},
    author={Wu, Sean and Amenta, Nicole and Zhou, Jiachen and Papais, Sandro and Kelly, Jonathan},
    booktitle={2023 20th Conference on Robots and Vision (CRV)},
    pages={89--96},
    year={2023},
    organization={IEEE}
}

@inproceedings{polley2024tld_ready,
    title={TLD-READY: Traffic Light Detection-Relevance Estimation and Deployment Analysis},
    author={Polley, Nikolai and Pavlitska, Svetlana and Boualili, Yacin and Rohrbeck, Patrick and Stiller, Paul and Bangaru, Ashok Kumar and Zollnerl, J Marius},
    booktitle={2024 IEEE 27th International Conference on Intelligent Transportation Systems (ITSC)},
    pages={3800--3806},
    year={2024},
    organization={IEEE}
}

@inproceedings{fan2025vitlr,
    author={Fan, Miao and Kong, Xuxu and Xu, Shengtong and Xiong, Haoyi and Liu, Xiangzeng},
    booktitle={2025 IEEE Intelligent Vehicles Symposium (IV)}, 
    title={Video-Based Traffic Light Recognition by Rockchip RV1126 for Autonomous Driving}, 
    year={2025},
    volume={},
    number={},
    pages={1738-1744},
    doi={10.1109/IV64158.2025.11097730}
}

@inproceedings{polley2025atlas,
    author={Polley, Rupert and Polley, Nikolai and Heid, Dominik and Heinrich, Marc and Ochs, Sven and Zöllner, J. Marius},
    booktitle={2025 IEEE Intelligent Vehicles Symposium (IV)}, 
    title={The ATLAS of Traffic Lights: A Reliable Perception Framework for Autonomous Driving}, 
    year={2025},
    volume={},
    number={},
    pages={2215-2222},
    doi={10.1109/IV64158.2025.11097347}
}

@article{chen2025yolo_ts,
    title={Yolo-ts: Real-time traffic sign detection with enhanced accuracy using optimized receptive fields and anchor-free fusion},
    author={Chen, Junzhou and Huang, Heqiang and Zhang, Ronghui and Lyu, Nengchao and Guo, Yanyong and Dai, Hong-Ning and Yan, Hong},
    journal={IEEE Transactions on Intelligent Transportation Systems},
    year={2025},
    publisher={IEEE}
}

@ARTICLE{lin2025yolo_llts,
  author={Lin, Ziyu and Wu, Yunfan and Ma, Yuhang and Chen, Junzhou and Zhang, Ronghui and Wu, Jiaming and Yin, Guodong and Lin, Liang},
  journal={IEEE Transactions on Instrumentation and Measurement}, 
  title={YOLO-LLTS: Real-Time Low-Light Traffic Sign Detection via Prior-Guided Enhancement and Multibranch Feature Interaction}, 
  year={2025},
  volume={74},
  number={},
  pages={1-18},
  doi={10.1109/TIM.2025.3604925}}

@inproceedings{wang2025rt_detr3,
    title={RT-DETRv3: Real-time end-to-end object detection with hierarchical dense positive supervision},
    author={Wang, Shuo and Xia, Chunlong and Lv, Feng and Shi, Yifeng},
    booktitle={2025 IEEE/CVF Winter Conference on Applications of Computer Vision (WACV)},
    pages={1628--1636},
    year={2025},
    organization={IEEE}
}

@misc{jocher2024yolo11,
    author = {Glenn Jocher and Jing Qiu},
    title = {Ultralytics YOLO11},
    version = {11.0.0},
    year = {2024},
    url = {https://github.com/ultralytics/ultralytics},
    orcid = {0000-0001-5950-6979, 0000-0003-3783-7069},
    license = {AGPL-3.0}
}

@inproceedings{lin2014coco,
    title={Microsoft coco: Common objects in context},
    author={Lin, Tsung-Yi and Maire, Michael and Belongie, Serge and Hays, James and Perona, Pietro and Ramanan, Deva and Doll{\'a}r, Piotr and Zitnick, C Lawrence},
    booktitle={European conference on computer vision},
    pages={740--755},
    year={2014},
    organization={Springer}
}

@misc{affinis2024trafficlightdetection,
    author       = {Affinis Lab},
    title        = {Traffic Light Detection Module},
    howpublished = {\url{https://github.com/affinis-lab/traffic-light-detection-module}},
    note         = {Module for detecting traffic lights in the CARLA simulator, based on YOLOv2 (Keras/TensorFlow).},
    year         = {2024}
}

@inproceedings{pavlitska2023tlr_survey,
    title={Traffic light recognition using convolutional neural networks: A survey},
    author={Pavlitska, Svetlana and Lambing, Nico and Bangaru, Ashok Kumar and Z{\"o}llner, J Marius},
    booktitle={2023 IEEE 26th International Conference on Intelligent Transportation Systems (ITSC)},
    pages={2790--2796},
    year={2023},
    organization={IEEE}
}

@inproceedings{zhao2024detrs,
    title={Detrs beat yolos on real-time object detection},
    author={Zhao, Yian and Lv, Wenyu and Xu, Shangliang and Wei, Jinman and Wang, Guanzhong and Dang, Qingqing and Liu, Yi and Chen, Jie},
    booktitle={Proceedings of the IEEE/CVF conference on computer vision and pattern recognition},
    pages={16965--16974},
    year={2024}
}

@inproceedings{redmon2016yolo,
    title={You only look once: Unified, real-time object detection},
    author={Redmon, Joseph and Divvala, Santosh and Girshick, Ross and Farhadi, Ali},
    booktitle={Proceedings of the IEEE conference on computer vision and pattern recognition},
    pages={779--788},
    year={2016}
}

@article{tang2025real,
    title={Real-time traffic light detection based on lightweight improved RT-DETR},
    author={Tang, Chaoli and Li, Yun and Wang, Lei and Li, Wenyan},
    journal={Journal of Real-Time Image Processing},
    volume={22},
    number={2},
    pages={1--15},
    year={2025},
    publisher={Springer}
}

@inproceedings{renz2025simlingo,
    title={Simlingo: Vision-only closed-loop autonomous driving with language-action alignment},
    author={Renz, Katrin and Chen, Long and Arani, Elahe and Sinavski, Oleg},
    booktitle={Proceedings of the Computer Vision and Pattern Recognition Conference},
    pages={11993--12003},
    year={2025}
}

@inproceedings{mei2024leapad,
    title={Continuously Learning, Adapting, and Improving: A Dual-Process Approach to Autonomous Driving},
    author={Jianbiao Mei and Yukai Ma and Xuemeng Yang and Licheng Wen and Xinyu Cai and Xin Li and Daocheng Fu and Bo Zhang and Pinlong Cai and Min Dou and Botian Shi and Liang He and Yong Liu and Yu Qiao},
    booktitle={The Thirty-eighth Annual Conference on Neural Information Processing Systems},
    year={2024}
}

@inproceedings{chen2024traffic,
    title={Traffic Light Detection and Recognition using Ensemble Learning with Color-Based Data Augmentation},
    author={Chen, Yong-Ci and Lin, Huei-Yung},
    booktitle={2024 IEEE Intelligent Vehicles Symposium (IV)},
    pages={3199--3204},
    year={2024},
    organization={IEEE}
}

@article{yu2022traffic,
    title={Traffic sign detection and recognition in multiimages using a fusion model with YOLO and VGG network},
    author={Yu, Jing and Ye, Xiaojun and Tu, Qiang},
    journal={IEEE Transactions on Intelligent Transportation Systems},
    volume={23},
    number={9},
    pages={16632--16642},
    year={2022},
    publisher={IEEE}
}

@article{wang2023vehicle,
    title={Vehicle-mounted adaptive traffic sign detector for small-sized signs in multiple working conditions},
    author={Wang, Junfan and Chen, Yi and Ji, Xiaoyue and Dong, Zhekang and Gao, Mingyu and Lai, Chun Sing},
    journal={IEEE Transactions on Intelligent Transportation Systems},
    volume={25},
    number={1},
    pages={710--724},
    year={2024},
    publisher={IEEE}
}

@article{langenberg2019deep,
    title={Deep metadata fusion for traffic light to lane assignment},
    author={Langenberg, Tristan and Lueddecke, Timo and Woergoetter, Florentin},
    journal={IEEE Robotics and Automation Letters},
    volume={4},
    number={2},
    pages={973--980},
    year={2019},
    publisher={IEEE}
}

@article{li2017traffic,
    title={Traffic light recognition for complex scene with fusion detections},
    author={Li, Xi and Ma, Huimin and Wang, Xiang and Zhang, Xiaoqin},
    journal={IEEE Transactions on Intelligent Transportation Systems},
    volume={19},
    number={1},
    pages={199--208},
    year={2017},
    publisher={IEEE}
}

@inproceedings{girshick2015fast,
    title={Fast r-cnn},
    author={Girshick, Ross},
    booktitle={Proceedings of the IEEE international conference on computer vision},
    pages={1440--1448},
    year={2015}
}

@article{redmon2018yolov3,
    title={Yolov3: An incremental improvement},
    author={Redmon, Joseph and Farhadi, Ali},
    journal={arXiv preprint arXiv:1804.02767},
    year={2018}
}

@article{simonyan2014vgg,
  title={Very deep convolutional networks for large-scale image recognition},
  author={Simonyan, Karen and Zisserman, Andrew},
  journal={arXiv preprint arXiv:1409.1556},
  year={2014}
}

@inproceedings{xing2025openemma,
    title={Openemma: Open-source multimodal model for end-to-end autonomous driving},
    author={Xing, Shuo and Qian, Chengyuan and Wang, Yuping and Hua, Hongyuan and Tian, Kexin and Zhou, Yang and Tu, Zhengzhong},
    booktitle={Proceedings of the Winter Conference on Applications of Computer Vision},
    pages={1001--1009},
    year={2025}
}

@article{sun2024llth,
    title={LLTH-YOLOv5: a real-time traffic sign detection algorithm for low-light scenes},
    author={Sun, Xiaoqiang and Liu, Kuankuan and Chen, Long and Cai, Yingfeng and Wang, Hai},
    journal={Automotive Innovation},
    volume={7},
    number={1},
    pages={121--137},
    year={2024},
    publisher={Springer}
}

@inproceedings{dosovitskiy2017carla,
    title = { {CARLA}: {An} Open Urban Driving Simulator},
    author = {Alexey Dosovitskiy and German Ros and Felipe Codevilla and Antonio Lopez and Vladlen Koltun},
    booktitle = {Proceedings of the 1st Annual Conference on Robot Learning},
    pages = {1--16},
    year = {2017}
}

@InProceedings{xie2025vlms,
        author    = {Xie, Shaoyuan and Kong, Lingdong and Dong, Yuhao and Sima, Chonghao and Zhang, Wenwei and Chen, Qi Alfred and Liu, Ziwei and Pan, Liang},
        title     = {Are VLMs Ready for Autonomous Driving? An Empirical Study from the Reliability, Data and Metric Perspectives},
        booktitle = {Proceedings of the IEEE/CVF International Conference on Computer Vision (ICCV)},
        month     = {October},
        year      = {2025},
        pages     = {6585-6597}
    }

@article{mao2023gpt,
    title={Gpt-driver: Learning to drive with gpt},
    author={Mao, Jiageng and Qian, Yuxi and Ye, Junjie and Zhao, Hang and Wang, Yue},
    journal={arXiv preprint arXiv:2310.01415},
    year={2023}
}

@inproceedings{wen2024dilu,
    title={DiLu: A Knowledge-Driven Approach to Autonomous Driving with Large Language Models},
    author={Licheng Wen and Daocheng Fu and Xin Li and Xinyu Cai and Tao MA and Pinlong Cai and Min Dou and Botian Shi and Liang He and Yu Qiao},
    booktitle={The Twelfth International Conference on Learning Representations},
    year={2024},
}

@ARTICLE{chen2025end,
    author={Chen, Li and Wu, Penghao and Chitta, Kashyap and Jaeger, Bernhard and Geiger, Andreas and Li, Hongyang},
    journal={IEEE Transactions on Pattern Analysis and Machine Intelligence}, 
    title={End-to-End Autonomous Driving: Challenges and Frontiers}, 
    year={2024},
    volume={46},
    number={12},
    pages={10164-10183},
    doi={10.1109/TPAMI.2024.3435937}
}

@article{yang2023llm4drive,
    title={Llm4drive: A survey of large language models for autonomous driving},
    author={Yang, Zhenjie and Jia, Xiaosong and Li, Hongyang and Yan, Junchi},
    journal={arXiv preprint arXiv:2311.01043},
    year={2023}
}

@article{guo2024drivemllm,
    title={Drivemllm: A benchmark for spatial understanding with multimodal large language models in autonomous driving},
    author={Guo, Xianda and Zhang, Ruijun and Duan, Yiqun and He, Yuhang and Zhang, Chenming and Liu, Shuai and Chen, Long},
    journal={arXiv preprint arXiv:2312.09245},
    year={2024}
}

@inproceedings{werby2024hierarchical,
  title={Hierarchical open-vocabulary 3d scene graphs for language-grounded robot navigation},
  author={Werby, Abdelrhman and Huang, Chenguang and B{\"u}chner, Martin and Valada, Abhinav and Burgard, Wolfram},
  booktitle={Robotics: Science and Systems (RSS)},
  year={2024}
}

@article{honerkamp2024language,
  title={Language-grounded dynamic scene graphs for interactive object search with mobile manipulation},
  author={Honerkamp, Daniel and B{\"u}chner, Martin and Despinoy, Fabien and Welschehold, Tim and Valada, Abhinav},
  journal={IEEE Robotics and Automation Letters},
  year={2024},
  publisher={IEEE}
}

@article{chisari2025robotic,
  title={Robotic Task Ambiguity Resolution via Natural Language Interaction},
  author={Chisari, Eugenio and von Hartz, Jan Ole and Despinoy, Fabien and Valada, Abhinav},
  journal={IEEE/RSJ International Conference on Intelligent Robots and Systems (IROS)},
  year={2025}
}

@article{mohammadi2025more,
  title={MORE: Mobile Manipulation Rearrangement Through Grounded Language Reasoning},
  author={Mohammadi, Mohammad and Honerkamp, Daniel and B{\"u}chner, Martin and Cassinelli, Matteo and Welschehold, Tim and Despinoy, Fabien and Gilitschenski, Igor and Valada, Abhinav},
  journal={IEEE/RSJ International Conference on Intelligent Robots and Systems (IROS)},
  year={2025}
}

\end{document}